\documentclass[10pt,onecolumn,letterpaper]{article}

\usepackage{cvpr}
\usepackage{times}
\usepackage{epsfig}
\usepackage{graphicx}
\usepackage{amsmath}
\usepackage{amssymb}
\usepackage{amsthm}
\usepackage{bm}
\usepackage[tight,footnotesize]{subfigure}
\usepackage{url}

\usepackage{algorithm}
\usepackage{algorithmic}

\usepackage{multirow}

\usepackage{eso-pic}
\usepackage{xspace}
\usepackage{array}
\newcolumntype{L}[1]{>{\raggedright\let\newline\\\arraybackslash\hspace{0pt}}m{#1}}
\newcolumntype{C}[1]{>{\centering\let\newline\\\arraybackslash\hspace{0pt}}m{#1}}
\newcolumntype{R}[1]{>{\raggedleft\let\newline\\\arraybackslash\hspace{0pt}}m{#1}}

\makeatletter
\DeclareRobustCommand\onedot{\futurelet\@let@token\@onedot}
\def\@onedot{\ifx\@let@token.\else.\null\fi\xspace}

\def\eg{\emph{e.g}\onedot} \def\Eg{\emph{E.g}\onedot}
\def\ie{\emph{i.e}\onedot} 
 
\def\etc{\emph{etc}\onedot} 
 
\def\etal{\emph{et al}\onedot}
\makeatother

\usepackage[pagebackref=true,breaklinks=true,letterpaper=true,colorlinks,bookmarks=false]{hyperref}

\cvprfinalcopy % *** Uncomment this line for the final submission

\begin{document}
% \renewcommand\thelinenumber{\color[rgb]{0.2,0.5,0.8}\normalfont\sffamily\scriptsize\arabic{linenumber}\color[rgb]{0,0,0}}
% \renewcommand\makeLineNumber {\hss\thelinenumber\ \hspace{6mm} \rlap{\hskip\textwidth\ \hspace{6.5mm}\thelinenumber}}
% \linenumbers

%%%%%%%%% TITLE
\title{Scalable Similarity Learning using Large Margin Neighborhood Embedding}

\author{
Zhaowen Wang\textsuperscript{\dag}, Jianchao Yang\textsuperscript{\ddag}, Zhe Lin\textsuperscript{\ddag},
Jonathan Brandt\textsuperscript{\ddag}, Shiyu Chang\textsuperscript{\dag}, Thomas Huang\textsuperscript{\dag} \\
\textsuperscript{\dag}Beckman Institute, University of Illinois at Urbana-Champaign, Urbana, IL 61801\\
\textsuperscript{\ddag}Adobe Research, San Jose, CA 95110\\
{\tt\small \{wang308, chang87, huang\}@ifp.uiuc.edu}, {\tt\small \{jiayang, zlin, jbrandt\}@adobe.com}
}

\maketitle
%\thispagestyle{empty}

%%%%%%%%% ABSTRACT
\begin{abstract}
Classifying large-scale image data into object categories is an important problem
that has received increasing research attention.
Given the huge amount of data, non-parametric approaches such as nearest neighbor classifiers
have shown promising results, especially when they are underpinned by a learned distance or similarity measurement.
Although metric learning has been well studied in the past decades, most existing algorithms are impractical to handle
large-scale data sets.
In this paper, we present an image similarity learning method that can scale well
in both the number of images and the dimensionality of image descriptors.
To this end, similarity comparison is restricted to each sample's local neighbors and
a discriminative similarity measure is induced from large margin neighborhood embedding.
We also exploit the ensemble of projections so that high-dimensional features
can be processed in a set of lower-dimensional subspaces in parallel without much performance compromise.
The similarity function is learned online using a stochastic gradient descent algorithm in which
the triplet sampling strategy is customized for quick convergence of classification performance.
The effectiveness of our proposed model is validated on several data sets with scales varying
from tens of thousands to one million images. Recognition accuracies competitive with
the state-of-the-art performance are achieved with much higher efficiency and scalability.
\end{abstract}

%%%%%%%%% BODY TEXT

\section{Introduction}

As the number of digital images generated and uploaded to the Internet skyrockets,
automatic categorization of large-scale image sets with diversified contents has become a popular research topic
\cite{Harchaoui12Trace,Liu13ProbTree,Li13Midlevel}.
The conventional approach to train a classifier for each class using one-versus-all paradigm
is usually unscalable to such a large number of images and classes,
not to mention that the sizes of most web data are constantly growing.
On the other hand, as the vast amount of image samples populate the data space more densely,
we can now afford classification models with higher capacities to capture the underlying data distribution.
Non-parametric methods, which infer the label information of test images from similar database images,
have demonstrated promising results on large data sets in many vision tasks including
scene parsing \cite{Tighe13}, object detection \cite{Malisiewicz11EgSVM}, face alignment \cite{Shen13Face}, \etc.
For classification, the non-parametric k-Nearest Neighbor (kNN) classifier \cite{Mensink12} has been successfully
applied on the ImageNet Challenge data set \cite{Deng09imagenet}.

A good distance or similarity measure is crucial to the performance of any non-parametric model.
As most image feature descriptors have very high dimensions and mainly characterize the low level visual information,
measuring their distance directly in the Euclidian space yields unsatisfactory results.
In order to close the semantic gap, people have used supervised information \cite{Xing02}
to learn distance metrics with promoted semantic similarity,
famous examples include information-theoretic \cite{Davis07} and
large margin nearest neighbor (LMNN) \cite{Weinberger09} methods.
The supervision is usually provided in the form of comparative constraints
over image pairs, triplets or even quadruplets \cite{Law13Quad},
leading to a time complexity which grows polynomially with sample size.
In addition, for large-scale data where multi-modal distributions are usually observed,
a single distance metric is insufficient to correctly measure the similarities
between all image pairs throughout the space.
To ameliorate the problem, multiple metrics have been applied to different parts of the data space,
by either assigning a distinct metric to each discrete space partition \cite{Hwang11,mu13}
or learning an adaptive metric parameterized based on the location of test
sample \cite{Domeniconi02,Noh10,Wang12parametric,Hauberg12Geo,Huang13Reduced}.
However, the extra model complexity of local metrics makes them less suitable for large-scale applications.

Besides the size of data sets, the high dimensionality of image descriptors
is another factor limiting the scalability of existing metric learning algorithms.
The Mahalanobis distance, one of the most popular forms of distance metric,
requires computation quadratic to data dimension in calculation
and cubic to dimension in its learning when positive-semidefinite constraint is placed.
Low-rank regularization can be imposed on the Mahalanobis metric to save computation,
but this may result in a non-convex optimization problem \cite{Mensink13} and cause performance loss in some cases.
Kernel function is useful in reducing the number of free metric parameters \cite{Wu12},
but it does not scale up well to the number of samples.
%it only works when we have a relatively small training set.

In this paper, we propose a new similarity learning algorithm that features good scalability
with respect to both sample size and dimensionality.
First, motivated by the findings from manifold learning with neighborhood embedding \cite{Yan07graph,Zhang08DNE},
we restrict similarity comparison to sample pairs within the same local neighborhood,
and try to capture the discriminative structure of local data manifold
using \emph{large margin neighborhood embedding} (Sec.~\ref{sec:slne}).
In this way, we can not only save a great amount of computation in training and testing,
but also gain robustness to outliers by focusing only on more relevant samples,
which shares the same rationale as the method in \cite{Yang06Efficient}.
On the other hand, we project the original high-dimensional data to a set of lower-dimensional subspaces,
and use the \emph{ensemble of similarities} learned from these subspaces as a surrogate to
the similarity in original space (Sec.~\ref{sec:slfe}).
The similarity for each subspace can be evaluated and optimized in parallel,
which offers superior scalability to data dimension.
The proposed method is validated on several image classification benchmarks with varying scales (Sec.~\ref{sec:exp}),
and both of its accuracy and efficiency are shown to scale up gracefully from tens of thousands to one million images.
Potential extensions of this work are also discussed (Sec.~\ref{sec:cncl}).
In short, the main contributions of this paper are:
\begin{itemize}
  \item a neighborhood embedding based similarity learning algorithm with improved classification performance and
        better scalability to the number of training samples;
  \item an ensemble of distributed similarities learning algorithm which scales well to data dimensionality.
\end{itemize}

%------------------------------------------------------------------------

\section{Similarity Learning using Neighborhood Embedding}
\label{sec:slne}

Graph embedding \cite{Yan07graph} is a family of dimensionality reduction algorithms which map data points
from a manifold in high-dimensional space to low-dimensional space while preserving the intrinsic data structure
represented by a weighted graph.
In many cases such as Locally Linear Embedding (LLE) \cite{Roweis00LLE} and Laplacian eigenmap \cite{Belkin03LEM},
a sparsely-connected graph is constructed based on neighborhood relationship, which we refer to as neighborhood embedding.

Since learning the transform of dimensionality reduction can be regarded as a special case of distance or similarity learning,
it is easy to extend the concept of neighborhood embedding to similarity learning.
Given a set of $N$ data samples $\{\mathbf{x}_i\}_{i=1}^N$ and the associated class labels $\{y_i\}_{i=1}^N$,
we define $\mathcal{N}_i$ as the index set for $\mathbf{x}_i$'s $k$ nearest neighbors (in Euclidean distance).
$\mathcal{N}_i$ can be divided into two mutually exclusive subsets $\mathcal{N}^+_i$ and $\mathcal{N}^-_i$,
%$\mathcal{N}_i = \mathcal{N}^+_i \bigcup \mathcal{N}^-_i$,
which denote the indices of $\mathbf{x}_i$'s neighbors with and not with label $y_i$, respectively.
An adjacency graph can be built in which each $\mathbf{x}_i$ is a vertex, and there is an undirected edge with weight $w_{ij}$
linking $\mathbf{x}_i$ and $\mathbf{x}_j$ if $i \in \mathcal{N}_j$ or $j \in \mathcal{N}_i$.
Generally, neighborhood embedding tries to find an optimal transform $f$ for all the data samples
by minimizing the following loss function:
\begin{equation}
    \mathcal{L} = \sum_{i \in \mathcal{N}_j \vee j \in \mathcal{N}_i} w_{ij} d\left(f(\mathbf{x}_i), f(\mathbf{x}_j) \right),
\end{equation}
where $f$ is usually a linear transform and $d(\cdot,\cdot)$ is the Euclidean distance measure in the transformed space.
For our purpose, we are interested in a good similarity function $s(\mathbf{x}_i, \mathbf{x}_j)$
defined as a mapping from a pair of samples $(\mathbf{x}_i, \mathbf{x}_j)$ to a real number that quantifies their semantic similarity.
Here we adopt a bilinear similarity function parameterized by a matrix $\mathbf{M}$:
\begin{equation}
\label{eq:sim}
    s_{\mathbf{M}}(\mathbf{x}_i, \mathbf{x}_j) = \mathbf{x}_i^T \mathbf{M} \mathbf{x}_j,
\end{equation}
and $\mathbf{M}$ is a symmetric matrix usually with a low rank constraint.
Bilinear similarity is commonly used in place of distance metric for its compactness \cite{Chechik10,Deng11hie}.
Its performance in many recognition tasks is found to be similar as the Mahalanobis distance,
which has better theoretical properties but requires a positive semidefinite parameter matrix.
With the neighborhood embedding formulation, the objective for similarity learning can be cast as
\begin{equation}
\label{eq:embed}
    \min_{\mathbf{M}} \mathcal{L}(\mathbf{M}) =
        \sum_{i \in \mathcal{N}_j \vee j \in \mathcal{N}_i} w_{ij} s_{\mathbf{M}}(\mathbf{x}_i, \mathbf{x}_j).
\end{equation}

We still need to define the weights $w_{ij}$, which encode the class label information.
Binary values ${\pm}1$ are commonly used to assign $w_{ij}$ based on whether
$\mathbf{x}_i$ and $\mathbf{x}_j$ come from the same class or not \cite{Zhang08DNE},
which essentially has the same effect as partitioning the adjacency graph into
a within-class graph and a between-class graph \cite{Yan07graph,Cai07,Chen05cvpr}
to impose a pairwise constraint that the similarity should be high between samples from the same class
and low otherwise.
For kNN classifiers, as suggested in \cite{Weinberger09}, we care more about the relative similarity
defined over a triplet of samples; \ie, a higher similarity score should be assigned when a sample $\mathbf{x}_i$
is compared with any of its target neighbor $\mathbf{x}_j, j{\in}\mathcal{N}^+_i \vee i{\in}\mathcal{N}^+_j$
than with any of its imposter neighbor $\mathbf{x}_l, l{\in}\mathcal{N}^-_i \vee i{\in}\mathcal{N}^-_l$.
To this end, we define the graph weights as
\begin{equation}
\label{eq:wij}
    w_{ij} = \left\{ \begin{array}{rl}
                       -|\{l| l{\in}\mathcal{N}^-_i \vee i{\in}\mathcal{N}^-_l \}|, & j{\in}\mathcal{N}^+_i \vee i{\in}\mathcal{N}^+_j \\
                       +|\{l| l{\in}\mathcal{N}^+_i \vee i{\in}\mathcal{N}^+_l \}|, & j{\in}\mathcal{N}^-_i \vee i{\in}\mathcal{N}^-_j \\
                       0, & \mathrm{otherwise}
                     \end{array}
     \right. ,
\end{equation}
where $|\mathcal{A}|$ denotes the cardinality of set $\mathcal{A}$.
With the weights in \eqref{eq:wij}, we can organize our objective function into a more interpretable form:
\begin{equation}
    \mathcal{L}(\mathbf{M}) = \sum_i \sum_{j \in \mathcal{N}^+_i \vee i \in \mathcal{N}^+_j} \sum_{l \in \mathcal{N}^-_i \vee i \in \mathcal{N}^-_l}
                        -s_{\mathbf{M}}(\mathbf{x}_i, \mathbf{x}_j)+s_{\mathbf{M}}(\mathbf{x}_i, \mathbf{x}_l) ,
\end{equation}
which enforces relative similarity constraint for each triplet $(\mathbf{x}_i, \mathbf{x}_j, \mathbf{x}_l)$ from the same neighborhood.
To apply more penalty to those triplets violating the constraint, we use a hinge function to promote
large margin of relative similarity difference, leading to the final form of our objective:
\begin{equation}
\label{eq:objlocalsim}
    \min_{\mathbf{M}} \mathcal{L}(\mathbf{M}) =
    \sum_i \sum_{j \in \mathcal{N}^+_i \vee i \in \mathcal{N}^+_j} \sum_{l \in \mathcal{N}^-_i \vee i \in \mathcal{N}^-_l}
    \left[b - s_{\mathbf{M}}(\mathbf{x}_i, \mathbf{x}_j) + s_{\mathbf{M}}(\mathbf{x}_i, \mathbf{x}_l) \right]_+ ,
\end{equation}
where $[\cdot]_+ = \max(\cdot, 0)$ is the hinge loss function, and $b>0$ is the minimum required margin
by which $\mathbf{x}_i$ should be more similar to a target neighbor $\mathbf{x}_j$
than to an imposter neighbor $\mathbf{x}_l$ as measured by $s_{\mathbf{M}}(\cdot, \cdot)$.

\subsection{Learning Algorithm}

Optimal linear transforms for neighborhood embedding can be found by solving a generalized eigen decomposition problem
using graph Laplacian. However, this approach is not applicable for large-scale data and the nonlinear objective
in Eq.~\eqref{eq:objlocalsim}.
Instead, we use an online learning method based on stochastic gradient descent \cite{Bottou10},
which is similar to \cite{Mensink12,Chechik10}.
Specifically, we iteratively go through the whole training set and randomly sample triplet
$\{\mathbf{x}_i, \mathbf{x}_j, \mathbf{x}_l\}$ which contributes non-zero cost to the objective in \eqref{eq:objlocalsim}.
The sub-gradient of the objective evaluated at the current triplet is then used to update the
parameter $\mathbf{M}$. The update is performed iteratively with a diminishing step size
and terminates upon convergence.

There can be a huge number of triplets to be considered in the objective function \eqref{eq:objlocalsim},
even though the similarity comparison is restricted to local neighbors.
Thus, a good sampling strategy to generate candidate triplets is essential to the speed of convergence
on large data sets.
For each training sample $\mathbf{x}_i$, we search for its
target neighbor $\mathbf{x}_j$ and imposter neighbor $\mathbf{x}_l$ according to
\begin{align}
\label{eq:sample_trip}
    \{j, l\} = &  \arg \max_{j', l'}
                    s_{\mathbf{M}}(\mathbf{x}_i, \mathbf{x}_{l'}) - s_{\mathbf{M}}(\mathbf{x}_i, \mathbf{x}_{j'}), \nonumber \\
               &    \mathrm{s.t.} \;\;\;\; j' \in \mathcal{N}^+_i \vee i \in \mathcal{N}^+_{j'}, \;\;
                                  l' \in \mathcal{N}^-_i \vee i \in \mathcal{N}^-_{l'} .
\end{align}
Optimizing \eqref{eq:sample_trip} returns $\mathbf{x}_i$'s most dissimilar target neighbor $\mathbf{x}_j$
and most similar imposter neighbor $\mathbf{x}_l$, so that the triplet violates the relative similarity constraint the most.
This can be solved efficiently with $|\mathcal{N}_i|$ similarity comparisons.

The overall procedure for Similarity Learning with Neighborhood Embedding (SL-NE) is summarized in Algorithm~\ref{alg:slne_train}.

\begin{algorithm}[t]
\caption{Similarity Learning with Neighborhood Embedding (SL-NE)}
\begin{algorithmic}[1]
    \REQUIRE labeled data set $\mathcal{S}=\{ \mathbf{x}_i \in \mathbb{R}^D, y_i \in \mathbb{Z}^+ \}$,
            required margin $b$, initial step size $\rho_0$
    \ENSURE similarity parameter $\mathbf{M}$
    \STATE initialize $\mathbf{M} \in \mathbb{R}^{D{\times}D}$ as identity matrix
    \STATE set $t=1$
    \WHILE{not converge}
        \STATE randomly permute data set $\mathcal{S}$
        \FOR{each $(\mathbf{x}_i, y_i) \in \mathcal{S}$}
            \STATE choose $(\mathbf{x}_j, \mathbf{x}_l)$ from constraint-violating pairs according to \eqref{eq:sample_trip}
            \STATE set step size $\rho={\rho_0}/{\sqrt{(t-1)/|{\mathcal{S}}| + 1}}$
            \STATE update $\mathbf{M} \leftarrow \mathbf{M} + \rho \cdot \mathbf{x}_i(\mathbf{x}_j-\mathbf{x}_l)^T$
            \STATE normalize the Frobenius norm of $\mathbf{M}$ and project to symmetric/low rank space (optional)
            \STATE $t \leftarrow t+1 $
        \ENDFOR
    \ENDWHILE
    \STATE return $\mathbf{M}$
\end{algorithmic}
\label{alg:slne_train}
\end{algorithm}

\subsection{Relation to LMNN}
\label{sec:lmnn}

The SL-NE introduced above has a similar objective function as the well-known LMNN method \cite{Weinberger09},
which is also based on triplet relative constraint.
The key difference is that in LMNN the $k$ nearest target samples of $\mathbf{x}_i$ are required to be more similar
to $\mathbf{x}_i$ than \emph{all the imposter samples in the global space}, which is an overly restrictive constraint
that even exceeds the condition for correct prediction with kNN classifier.
As noted in \cite{Yang06Efficient,mu13}, such strong global constraints often conflict with each other for high-dimensional data
with multi-modal distribution, and makes the learning result ineffective and more vulnerable to outliers.
From another perspective, it is also possible to formulate LMNN using the graph embedding framework as in Eq.~\eqref{eq:embed},
and the associated adjacency graph will be densely connected due to many non-zero weights $w_{ij}$,
a stark contrast to our sparse graph whose edge connections are restricted inside local neighborhoods.
It is widely concurred that sparse graphs are typically superior to or more robust than dense graphs \cite{Zhu06semi}.
The proposed SL-NE method focuses on the similarity relationship among neighbors, and therefore is
more consistent with the requirement of kNN classifier and learns a more robust similarity function from
those relevant constraints.

\begin{figure}[t]
\center
\begin{tabular}{cc}
    \includegraphics[height=45mm,angle=0,clip]{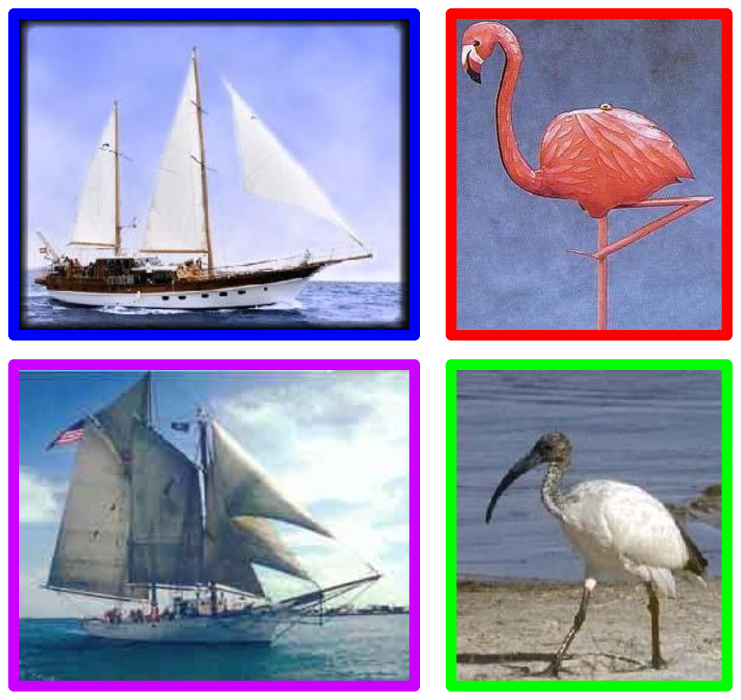} &
    \includegraphics[height=45mm,angle=0,clip]{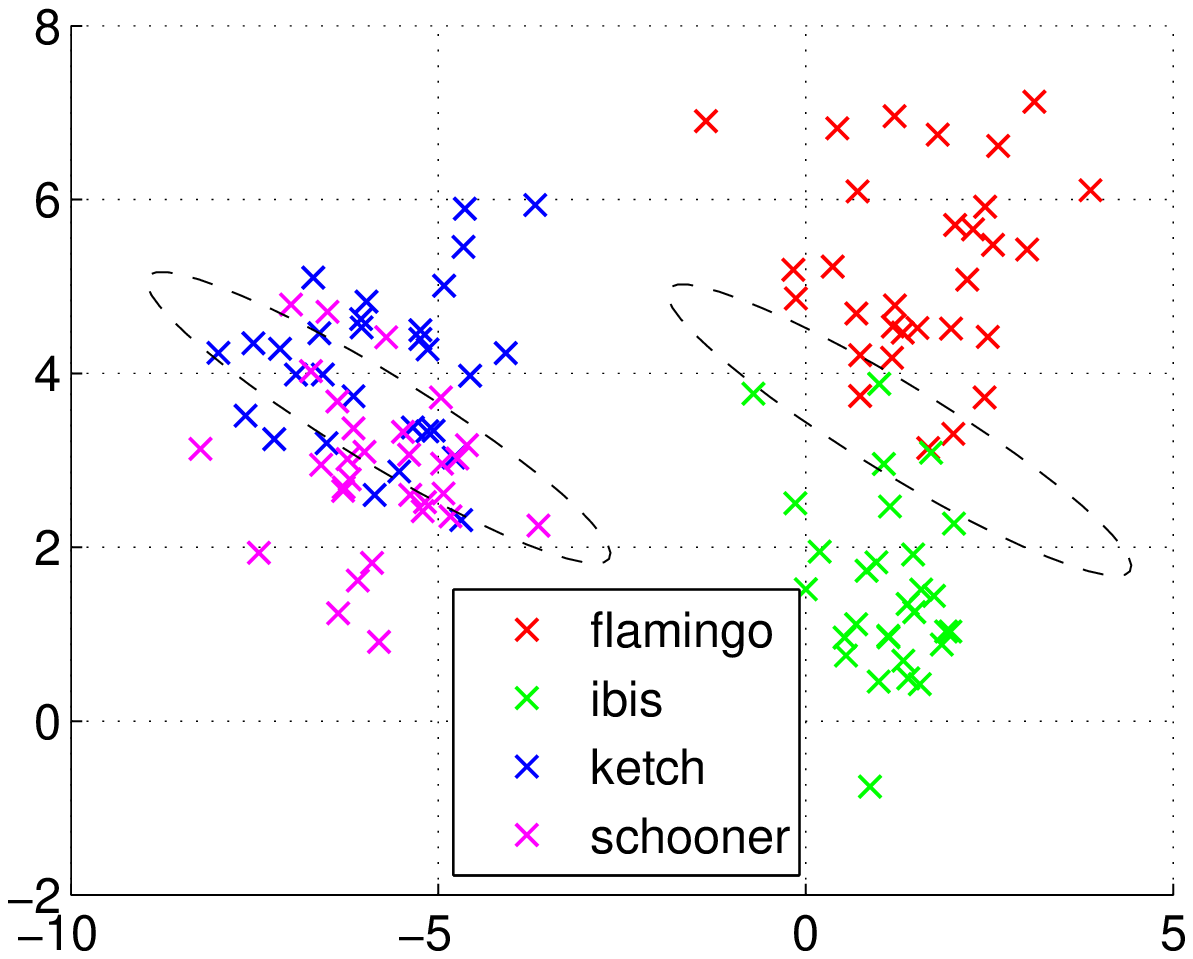} \\
    (a) & (b)
\end{tabular}
\caption{Four classes selected from Caltech 101 with image samples shown in (a) and distributions in 2D LDA space plotted in (b).
         The ellipses plotted at the two cluster centers represent the equal-similar contour learned from SL-NE.}
\label{fig:toy_cal101}
\end{figure}

\begin{table}[t]
\centering
\caption{Classification accuracies (\%) on the 4-class subset of Caltech 101 by different class divisions.}
\label{tab:toy_cal101}
\begin{tabular}{R{2.5cm}||C{1.5cm}|C{1.5cm}|C{1.5cm}|C{1.5cm}}
\hline
method	    & bird/boat & flm/ibs & ktc/sch & all four \\
\hline
Euclidean kNN	& 92.15 & 60.93 & 59.52 & 55.45  \\
LMNN        & 94.00 & 64.93 & 66.66 & 61.83 \\
SL-NE       & 98.72 & 78.14 & 76.21 & 76.43 \\
\hline
\end{tabular}
\end{table}

More concretely, the advantage of SL-NE over LMNN can be illustrated through an example shown in Fig.~\ref{fig:toy_cal101}
and the corresponding accuracy comparisons listed in Table~\ref{tab:toy_cal101}.
LMNN can discriminate well between two coarse-grained classes ``bird'' and ``boat'',
but fails to learn the subtle differences between fine-grained categories, \ie ``flamingo'' vs ``ibis''
or ``ketch'' vs ``schooner''. This is because a large part of LMNN's effort is wasted on
optimizing unimportant similarity constraints (\eg ``ibis'' vs ``ketch''),
which makes the learning less effective.
On the other hand, SL-NE finds the local discriminative structure due to its neighborhood embedding formulation,
and achieves much higher accuracies than both kNN with Euclidean distance and LMNN.
It is noted that metrics organized in a tree structure \cite{Hwang11,Verma12} have been proposed
to learn similarities with varying granularity in an object category hierarchy.
SL-NE can pick the most discriminative granularity level and learn a similarity function
without any knowledge on object ontology.

The philosophy behind SL-NE is to learn a shared local data structure with
information from different neighborhoods complementing each other,
which is also used in many other machine learning models.
\Eg, in Gaussian mixture model and Relevant Component Analysis (RCA) \cite{Bar03},
the covariance matrices are tied for all the mixture components or classes;
in the localized versions of Neighborhood Component Analysis (NCA) \cite{Yang06Efficient,Yang12fast}
and Fisher Discriminant Analysis (FDA) \cite{Sugiyama07}, the importance of a training pair
is weighted according to their affinity.
It should be noted that in the NCA methods \cite{Yang06Efficient,Yang12fast}, the neighborhood is
dynamically updated according to current distance metric. While in SL-NE, the neighborhood is fixed
throughout the learning iteration (with neighbors' ranks being updated),
which is more scalable to large data sets.

Besides the performance advantage, it is obvious to see the computational saving of SL-NE over LMNN.
The total number of triplet constraints to be considered is on the order of $N^2 k$ for LMNN which
regards all the samples with labels other than $y_i$ as the imposters for $\mathbf{x}_i$,
while only around $N k^2$ for SL-NE which defines similarity exclusively for neighboring samples.
Therefore, a significant reduction of computational complexity by a factor of $N/k$ can be achieved
for large-scale data with $N{\gg}k$.

%------------------------------------------------------------------------

\section{Ensemble of Distributed Similarities}
\label{sec:slfe}

Recent progress in image classification has witnessed the benefits of building features
with very high dimensionality \cite{Perronnin10}.
However, metric learning on such high-dimensional features can be prohibitively expensive.
Many methods have tried to learn the metric on a subspace with much lower dimension
by imposing a low rank constraint on the distance kernel \cite{Weinberger09,Mensink12},
which, unfortunately, makes the problem non-convex \cite{Mensink13}.
Although the bilinear similarity function in \eqref{eq:sim} takes a very simple form, it still has computation complexity
quadratic to data dimension.
To make our method also scalable to high-dimensional data in terms of computation and data I/O,
we propose to use an ensemble of low-dimensional subspace projections so that the learning
and evaluation of the similarity function can be conducted in a set of low-dimensional spaces distributedly.
Specifically, based on the similarity in \eqref{eq:sim}, we define an ensemble of similarities as
\begin{equation}
\label{eq:ensmbl_sim}
          s_E(\mathbf{x}_i, \mathbf{x}_j)
      =  \sum\limits_{n=1}^{N_E} s_{\mathbf{M}_n}(\mathbf{P}_n \mathbf{x}_i, \mathbf{P}_n \mathbf{x}_j)
      =  \sum\limits_{n=1}^{N_E} \mathbf{x}_i^T \mathbf{P}^T_n \mathbf{M}_n \mathbf{P}_n \mathbf{x}_j ,
     % =  \mathbf{x}_i^T \sum\limits_{n=1}^{N_E} \left( \mathbf{P}^T_n \mathbf{M}_n \mathbf{P}_n \right) \mathbf{x}_j,
\end{equation}
where $\{\mathbf{P}_n\}_{n=1}^{N_E}$ is a set of $d{\times}D$ matrices that project data samples $\{\mathbf{x}_i\}$
from the original $\mathbb{R}^D$ space to $\mathbb{R}^d$ spaces, with $d < D$.
$N_E$ is the number of projections used in the ensemble.
$\mathbf{M}_n$ is the parameter of the similarity function in the $n$-th projected space.
Ideally, $\{\mathbf{P}_n\}_{n=1}^{N_E}$ should be a set of projections capturing complementary discriminative information.
However, learning these discriminative projections can be expensive and thus spoils the algorithm's scalability.
In practice, we find that projections built as consecutive partitions of PCA directions
%\footnote{obtained by evenly partitioning PCA dimensions into consecutive and disjoint blocks in the order of eigenvalues.}
and random projections are both good candidates for $\{\mathbf{P}_n\}$. With the additional benefit of knowing the energy in each
projection, PCA projections decorrelate the data in different subspaces, which guarantees their complementarity in certain sense.
PCA directions can be efficiently approximated using a subset of data, and it entails a one-time computation
as opposed to low-rank metric which performs extra calculation in each training iteration,
Random projections also have several attractive properties. First, they can be obtained virtually at no cost.
Besides, they can well preserve distance in high-dimensional space according to the Johnson-Lindenstrauss lemma \cite{Johnson84},
as well as low-rank data structure according to \cite{Halko11}.

From \eqref{eq:ensmbl_sim}, we can see that learning the similarity function $s_{\mathbf{M}_n}$ in
the $n$-th projected space is equivalent to learning a similarity function $s_{\mathbf{M}}$ in the original
space, with low rank and subspace constraints induced by $\mathbf{P}_n$ and imposed on $\mathbf{M}$.
From \eqref{eq:ensmbl_sim}, we can further interpret that the ensemble of similarity
functions tries to approximate the complete space parameter $\mathbf{M}$ with the summation of
a set of parameters $\{\mathbf{M}_n\}$ constrained to subspaces.
%Therefore, with sufficiently large $N$, the ensemble of $N$ similarity functions learned in their projected low-dimensional
%spaces have the same capacity as the similarity function learned in the original high dimensional space.

Ensemble learning of multiple metrics has been explored in literatures with different settings.
A local distance metric method is proposed in \cite{mu13}, which learns a metric for each training sample and combines
them in the form of class probability prediction. Such an approach is not scalable to large data sets.
Boosting algorithms are employed to select and combine multiple weak metrics in \cite{Huang13,Kedem12}.
The metrics have to be learned sequentially, which is not efficient when the number of metrics is large.
The ensemble method proposed here focuses on parallel learning of multiple metrics.
Such computational advantage is also leveraged by the random forest metric \cite{Xiong12}, which regresses the distance function
as the average of binary outputs from a set of decision trees.
Another method of ensemble metric learning with parallel capability is introduced in \cite{Kozakaya11},
%where the metric is constructed by concatenating the coefficients of Support Vector Machines (SVM)
%trained on randomly subsampled data sets.
where the ensemble is based on different partitions of class subsets so that better scalability to
sample size instead of sample dimension is achieved.

\subsection{Learning Algorithm}

%Instead of attempting to jointly learn all the parameters $\{\mathbf{M}_n, \mathbf{w}_n\}$,
%we adopt a more efficient parallel approach to learn each low-dimensional similarity function
%$s_{\mathbf{M}_n, \mathbf{w}_n}$ independently in parallel.
%The factorization of such a large optimization problem into several independent smaller ones is plausible
%if the data projected in different random subspaces are somehow uncorrelated.

Learning the ensemble similarity $s_E(\cdot, \cdot)$ requires optimizing the objective function
in \eqref{eq:objlocalsim} with the similarity defined in \eqref{eq:ensmbl_sim}.
Applying SL-NE in Algorithm~\ref{alg:slne_train} to find all the $\mathbf{M}_n$'s jointly
induces a computational complexity of $O(N_E{\cdot}d^2)$,
which may not offer too much gain over the original complexity of $O(D^2)$
if a large ensemble size $N_E$ is used.
Instead, we propose a Similarity Learning with Distributed Ensemble (SL-DE) algorithm
with the computational advantage that each single similarity function in the ensemble
can be learned independently in a distributed manner.
%Just like random forest \cite{Breiman01}, the proposed
Given the projection matrix $\mathbf{P}_n$, each $\mathbf{M}_n$ is learned \emph{in parallel}
in the projected space of $\mathbf{P}_n \mathbf{x}$ using SL-NE,
which has time complexity $O(d^2)$.
The resulting similarity functions can be directly combined to approximate
the optimal ensemble similarity $s_E(\cdot, \cdot)$ according to \eqref{eq:ensmbl_sim}.
In this way, we can potentially reduce the computational complexity from $O(D^2)$ to $O(d^2)$
if not considering the overhead in parallelization.
Learning the similarity functions in low-dimensional spaces also helps the optimization converge more quickly,
which offers additional saving in computation.

Since the individually learned $\mathbf{M}_n$'s are suboptimal, when computation resource allows,
we can further carry out an optional step that jointly optimizes them
by minimizing the objective in \eqref{eq:objlocalsim}.
The joint optimization can be done in a coordinate descent manner,
where each $\mathbf{M}_n$ is sequentially updated with all the others in $s_E(\cdot, \cdot)$ fixed.
With a reasonable initialization from the individually trained $\mathbf{M}_n$'s,
this joint optimization does not take long to converge in practice.

Lastly, we want to point out that the idea of SL-DE can be used to accelerate other general metric learning methods
and is not limited to SL-NE.

%------------------------------------------------------------------------

\section{Experiments}
\label{sec:exp}

%\subsection{Implementations and Settings}

In this section, we test the performance of the proposed SL-NE learning algorithm and its ensemble version SL-DE
on several data sets.

In all the experiments, we use Locality-constrained Linear Coding (LLC) \cite{Wang10LLC} as the image feature.
With code book size $2048$ and spatial pyramid matching on the $1{\times}1$, $2{\times}2$ and $4{\times}4$ grids,
the LLC feature representation is of $43{,}008$ dimensions.
Unless otherwise specified, we project the LLC features to the $3{,}000$ dimensional PCA space
and normalize their lengths, which is regarded as the input feature space
for all the metric learning methods considered here.
For the SL-NE and SL-DE methods, we set margin $b{=}0.02$ and initial step size $\rho_0{=}0.2$.%, and $20$ total iterations.
The training of SL-DE is performed on a number of distributed nodes in a computer cluster.
The local neighborhood $\mathcal{N}_i$ is approximately found using an image retrieval
system \cite{Shen12}\footnote{We are grateful to the authors for providing the executable.}
which is efficient for large-scale applications and has a high recall of images from the same class.
Note that SL-NE is open to any neighborhood construction method, including
approximate nearest neighbor search and hashing.
The same neighborhood found in the original feature space is used to learn
the similarity of each projected subspace in SL-DE.

Our learned similarity functions are used with a soft-voting kNN classifier to make prediction for test samples,
where the voting weight are given by the similarity scores. Specifically, for a test sample $\mathbf{x}_t$
whose $k$ nearest training neighbors are indexed by a set $\mathcal{N}_t$, its class label can be predicted as:
\begin{equation}
    \hat{y}_t = \arg \max_c \sum\limits_{j \in \mathcal{N}_t, y_j=c} s(\mathbf{x}_t, \mathbf{x}_j) .
\end{equation}
The voting members for each test sample are selected from its local neighborhood instead of
the entire training set so that the testing is also scalable to sample size.

Our SL-NE and SL-DE methods are compared with linear SVM, the Retrieval system \cite{Shen12} which generates the
initial neighborhoods, and LMNN \cite{Weinberger09} whose code is publicly available.
A rough comparison on training complexity for all these five methods is given in Table \ref{tab:timecomplex}.
%SVM uses one-versus-all approach for multi-class classification, and is not scalable to the number of classes.
%LMNN needs to consider similarity constraints with order quadratic to the sample number
Generally, metric learning methods are more scalable than one-versus-all SVM in terms of the number of classes.
Our SL-NE is more scalable than LMNN in terms of the number of training samples due to its neighborhood embedding formulation,
and SL-DE further improves over SL-NE on the scalability of data dimension by parallel computation.

\begin{table}[t]
\centering
\caption{Training time complexities.}
\label{tab:timecomplex}
\begin{tabular}{r||C{2.5cm}C{2.5cm}C{3cm}}
\hline
	& sample number    & class number    & data dimension\\
\hline
linear SVM	& linear	& linear	& linear \\
Retrieval \cite{Shen12}	& linear	& constant	& constant \\
LMNN \cite{Weinberger09}	& quadratic	& constant	& quadratic \\
SL-NE	& linear	& constant	& quadratic \\
SL-DE	& linear	& constant & constant (parallel) \\
\hline
\end{tabular}
%\vspace{-3mm}
\end{table}

In the following, we first analyze in Sec.~\ref{sec:exp_anly} the characteristics of SL-NE and SL-DE algorithms
on the small-sized Caltech 101 data set \cite{Fei07}.
Then more results of classification on middle-scale and large-scale data are discussed in
Sec.~\ref{sec:exp_mid} and \ref{sec:exp_lrg}, respectively.

\subsection{Algorithm Analysis}
\label{sec:exp_anly}

\begin{figure}[t]
\center
    \includegraphics[width=70mm,angle=0,clip]{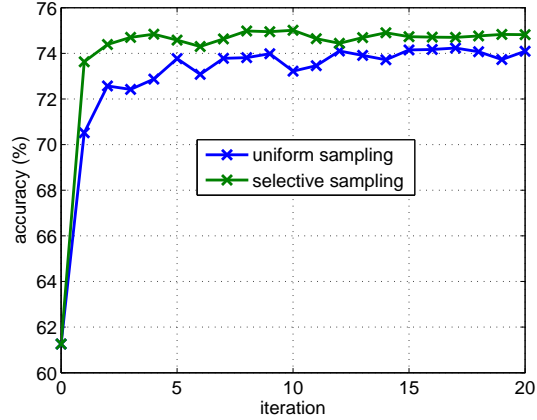}
\caption{Accuracy on test set versus training iterations on the Caltech 101 data using SL-NE.}
%\vspace{-3mm}
\label{fig:itr_cal101}
\end{figure}

\begin{figure}[t]
\center
    \begin{tabular}{cc}
        \includegraphics[width=75mm,angle=0,clip]{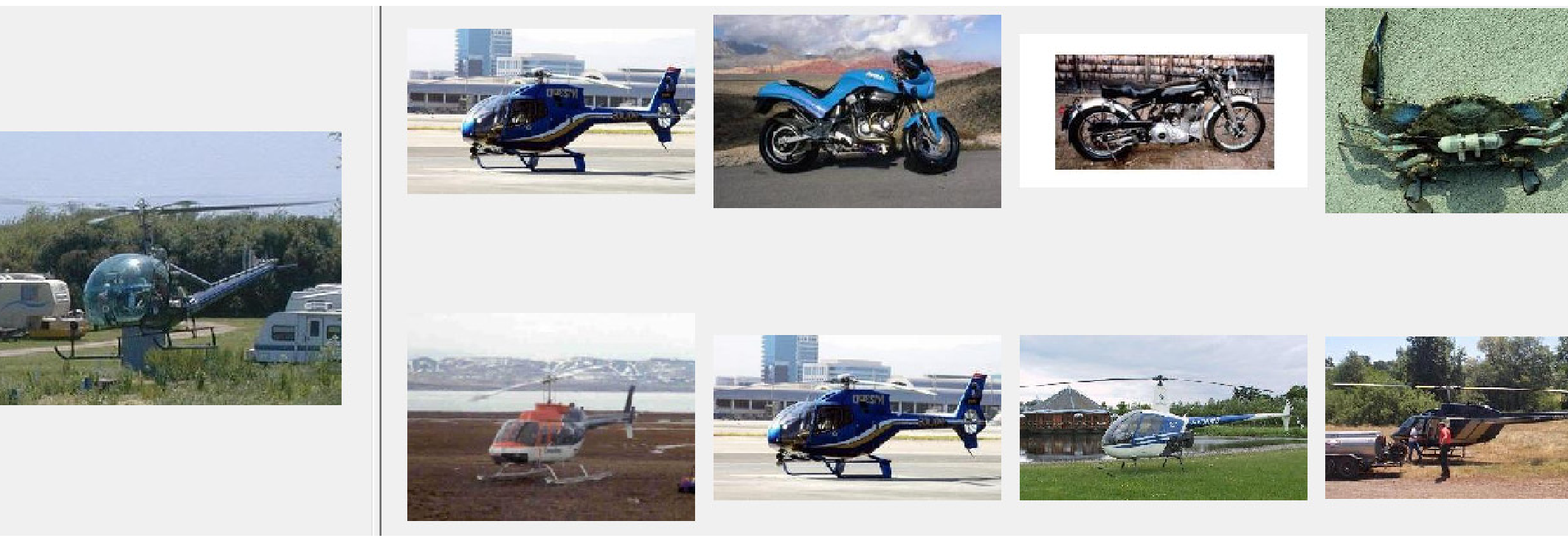} &
        \includegraphics[width=75mm,angle=0,clip]{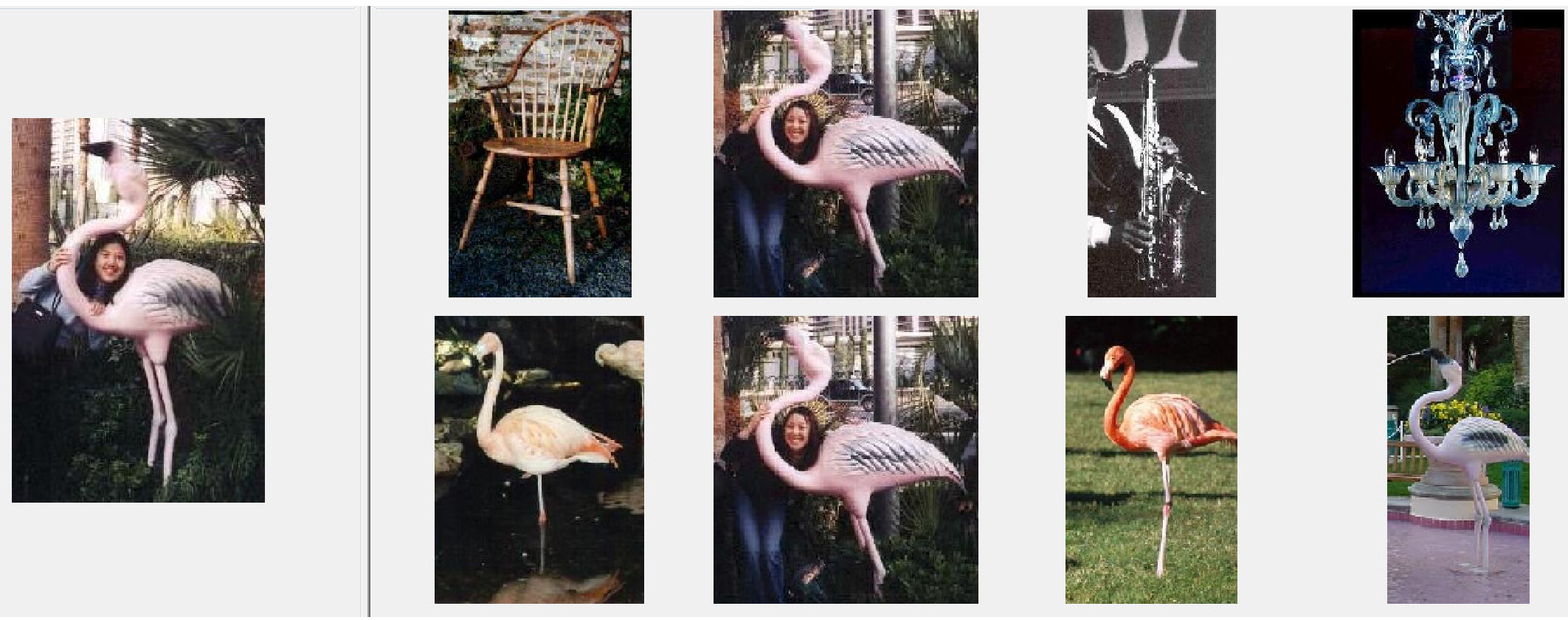} \\
         (a) & (b) \\
        \includegraphics[width=75mm,angle=0,clip]{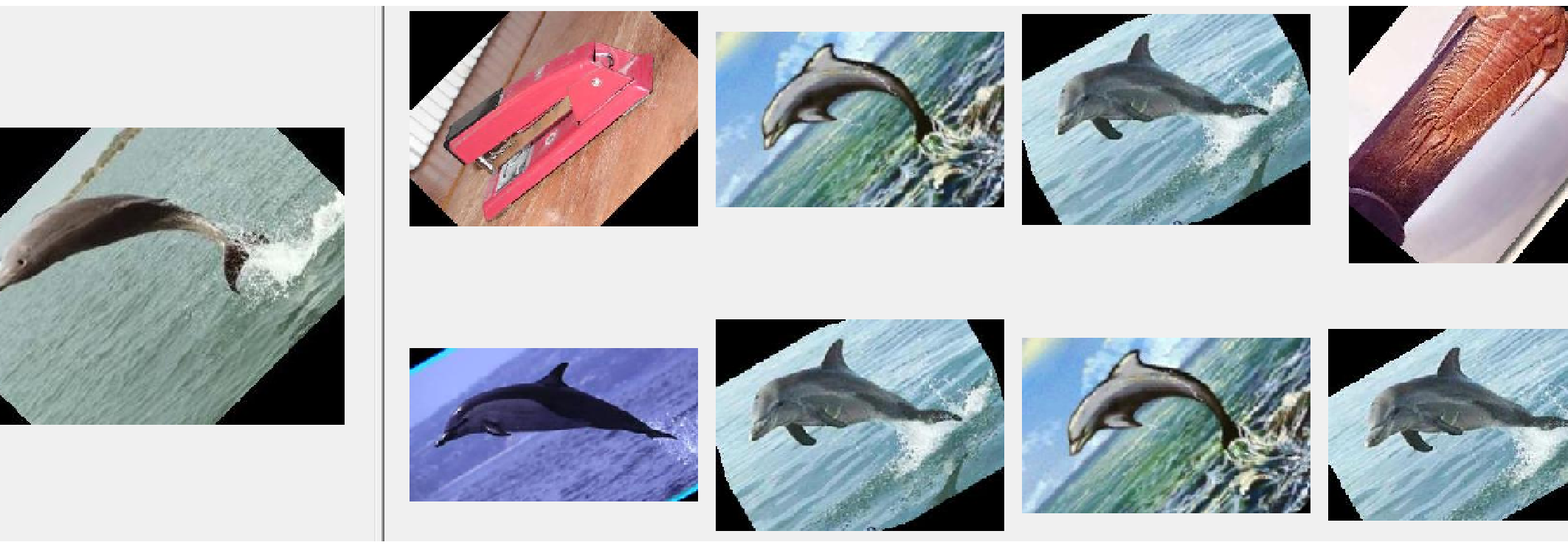} &
        \includegraphics[width=75mm,angle=0,clip]{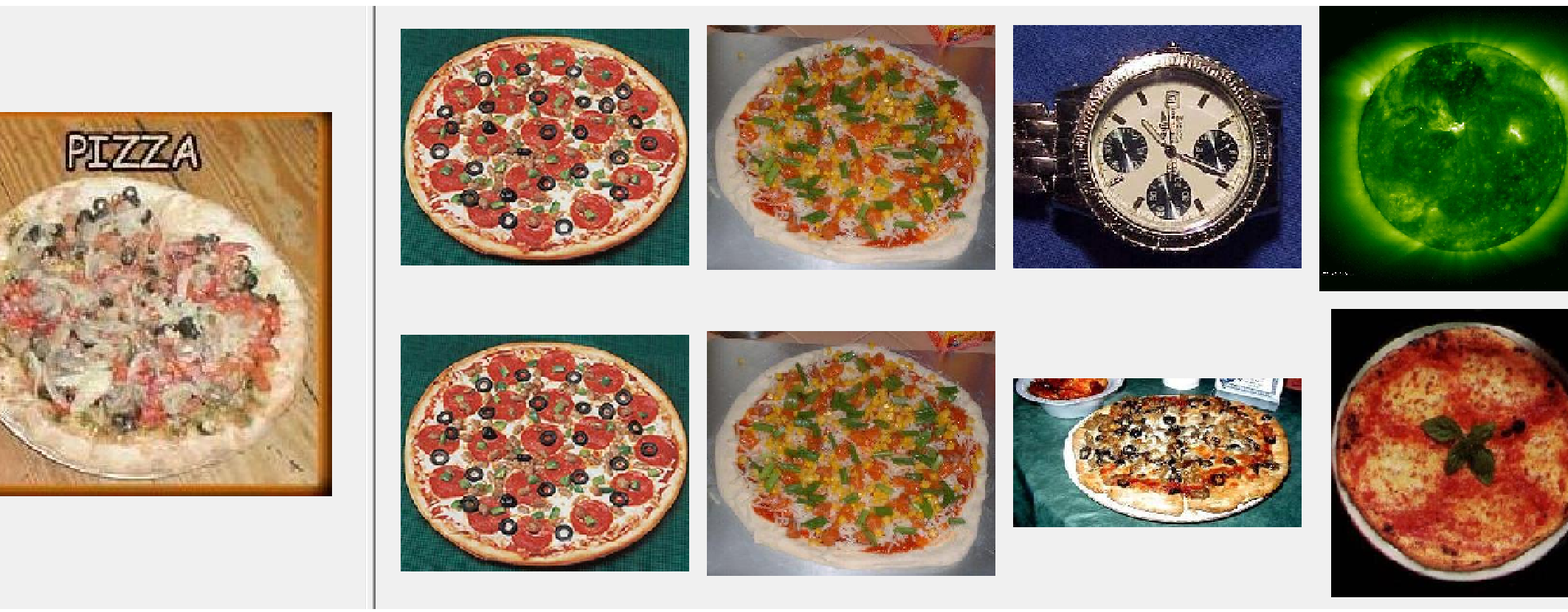} \\
         (c) & (d)
    \end{tabular}
\caption{(a)$\sim$(d): A test image is shown in the left, and the training images most similar to it is shown in the right,
             found by Retrieval \cite{Shen12} (top row) and SL-NE (bottom row).}
\label{fig:retr_cal101}
\end{figure}

We first examine the optimization behavior for SL-NE algorithm in Fig.~\ref{fig:itr_cal101}.
Two types of triplet sampling strategy are compared: uniform sampling and the selective sampling defined by \eqref{eq:sample_trip}.
With either sampling strategy, the performance of the learned similarity function improves a lot over the initial inner product
similarity in the original space. Moreover, the proposed selective sampling can converge faster than uniform sampling and attains higher accuracy.
Some examples of similar images found by Retrieval and the SL-NE algorithm are shown in Fig.~\ref{fig:retr_cal101}.
It can be seen that the images found by SL-NE are more semantically similar to the test images and are more likely to come from the
same classes.

\begin{figure}[!t]
\center
    \begin{tabular}{cc}
        \includegraphics[height=60mm,angle=0,clip]{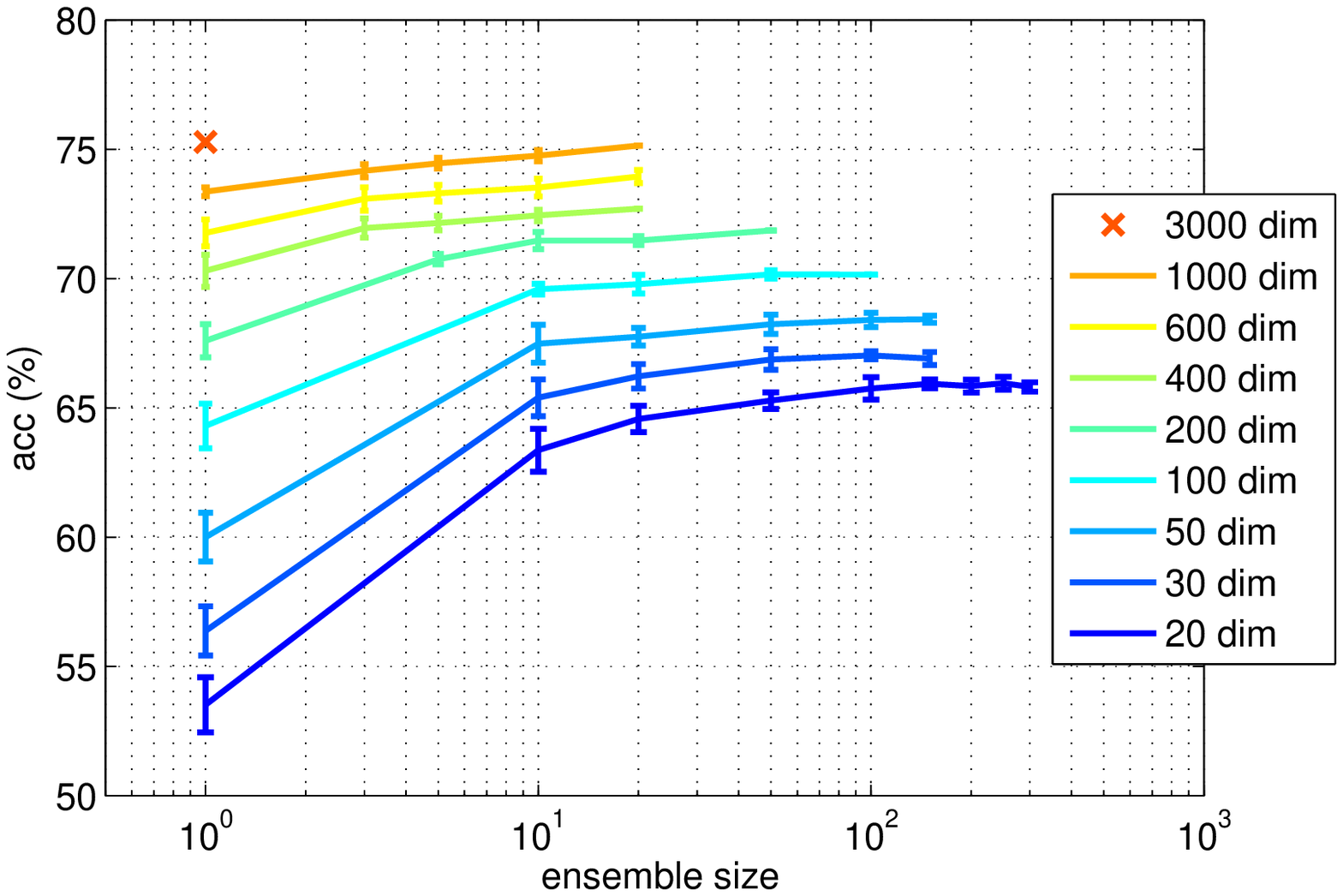} &
        \includegraphics[height=60mm,angle=0,clip]{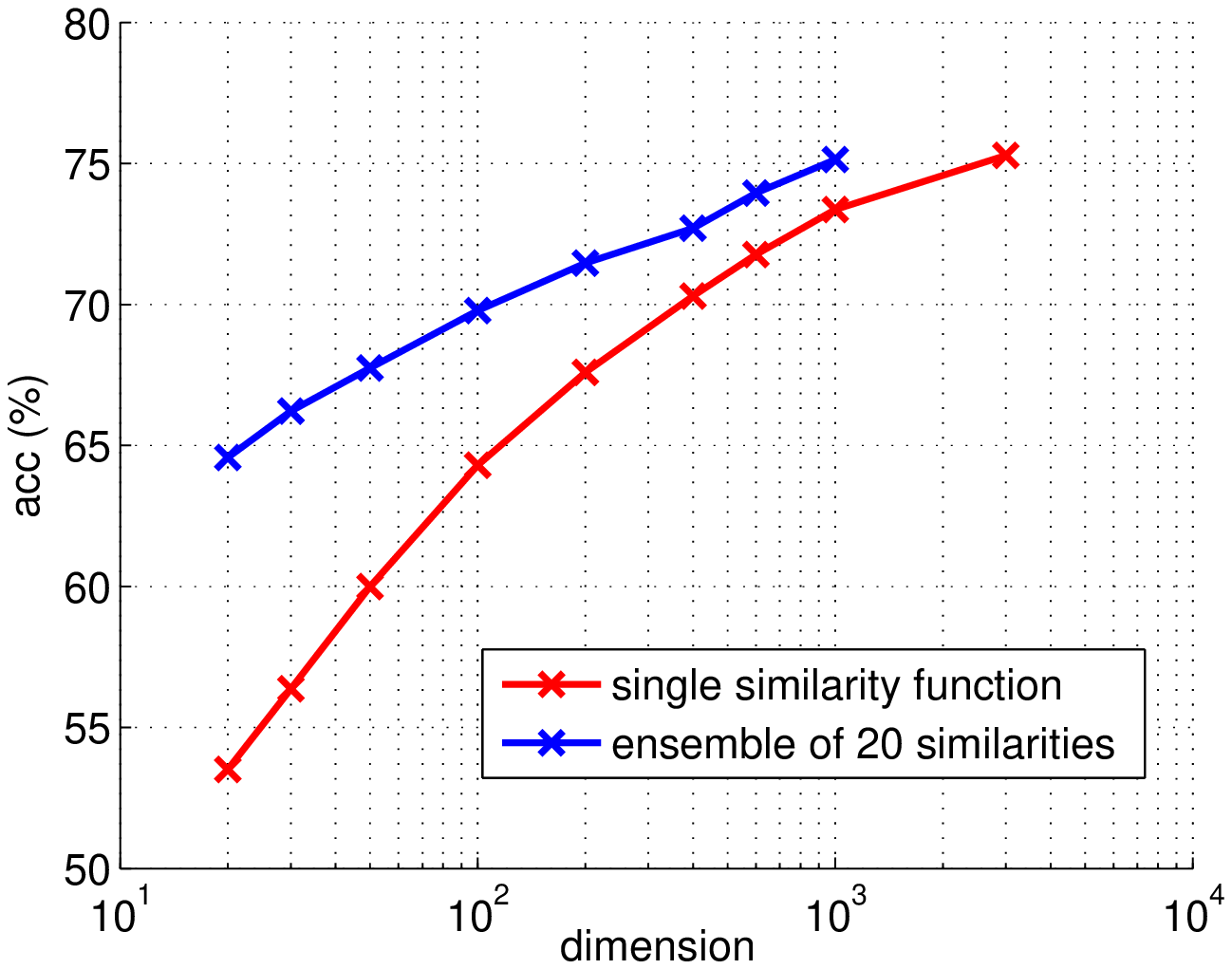}  \\
         (a) & (b)
    \end{tabular}
\caption{Test accuracy of SL-DE using random projections on the Caltech 101 data:
         (a) plotted as a function of ensemble size $N_E$ for various projected dimension $d$'s;
         (b) plotted as a function of projected dimension $d$ for $N_E=1$ and $N_E=20$ similarity function(s) used in ensemble.}
\vspace{-3mm}
\label{fig:ensemble_size_cal101}
\end{figure}

The performance of SL-DE is studied in Fig.~\ref{fig:ensemble_size_cal101} with different
combinations of projection dimension $d$ and ensemble size $N_E$.
Here we project image features from the original $3{,}000$ dimensional space
to random subspaces of dimensions ranging from $20$ to $1{,}000$.
From Fig.~\ref{fig:ensemble_size_cal101} (a), we see that the accuracy of SL-DE increases with the size of ensemble,
and converges to a certain bound determined by the projection dimension. An ensemble of $10$ similarity functions
in $1{,}000$ dimensional projected spaces, for example, can achieve almost the same performance as
the single similarity function learned in the original $3{,}000$ dimensional space.
The total computation for the two are about the same\footnote{$10{\times}1{,}000^2\approx3{,}000^2$},
but the ensemble approach can be more easily parallelized.
It is further observed from (b) that, compared to the single similarity function learned by SL-NE,
SL-DE with an ensemble size $20$ can work in subspaces of much reduced dimension without compromise in performance.

\subsection{Results on Middle Scale Data}
\label{sec:exp_mid}

The classification performance of our methods are validated on several benchmark data sets including
the Caltech 101 \cite{Fei07} (9{,}144 images from 102 object classes),
the Caltech 256 \cite{Griffin07} (30{,}607 images from 257 object classes)
and the SUN \cite{Xiao10} (108{,}754 images from 397 scene categories).
We randomly select 30 samples/80 samples/70\% samples from each class as training set
for the Caltech 101/Caltech 256/SUN data set, respectively.
In our SL-NE and SL-DE methods, the local neighborhood size $|\mathcal{N}|$ is chosen as 50/50/500
in both training and testing.
The neighborhood size is selected to be about the same as the size of each class in the data sets.
For the SL-DE method, an ensemble of similarity functions in 1{,}000/300/500 dimensional PCA and random subspaces
is trained for the three data sets respectively.
Note we just choose the ensemble parameters arbitrarily as long as the computation resource allows.
The top-1 and top-3 classification accuracies are shown in Table~\ref{tab:acc}, with comparison to several baseline approaches.
On all the data sets, our SL-NE method is much better than the unsupervised Retrieval,
and most of the time it also outperforms the popular linear SVM classifier by 2$\sim$3\%.
SL-NE also achieves much higher accuracies than LMNN (which cannot complete in a reasonable amount of time on the SUN set).
This indicates that the neighborhood embedding formulation can not only reduce training complexity, but also improve
the learning effectiveness by focusing on more relevant data.
The accuracies attained by the SL-DE method are very close to those of SL-NE,
even though SL-DE are learned based on features projected to subspaces with much lower dimensions.
On the SUN data set, SL-DE even performs better than SL-NE, which implies that sometimes directly learning a similarity function
in high-dimensional space may not be as effective as the ensemble approach.

\begin{table}[t]
\centering
\caption{Top $n$ classification accuracies (\%) for middle scale data sets.}
\label{tab:acc}
\begin{tabular}{r||c|C{1.5cm}C{2cm}C{2cm}C{1.5cm}C{1.5cm}}
\hline
Data set	                    & Top-$n$	& SVM	& Retrieval \cite{Shen12}	& LMNN \cite{Weinberger09}	& SL-NE	& SL-DE \\
\hline
\multirow{2}{*}{Caltech 101}	& 1	& 73.46	& 63.49	& 69.16	& \textbf{75.28}	& 75.14 \\
                            	& 3	& 84.59	& 77.87	& 80.83	& 85.29	& \textbf{85.30} \\
\hline
\multirow{2}{*}{Caltech 256}	& 1	& 43.61	& 37.28	& 37.47	& \textbf{46.02}	& 45.39 \\
	                            & 3	& \textbf{56.97}	& 48.83	& 46.84	& 55.97	& 55.33 \\
\hline
\multirow{2}{*}{SUN}	    & 1	& 32.08	& 29.06	& --	& 35.32	& \textbf{35.72} \\
	                            & 3	& 48.73	& 43.87	& --	& 51.37	& \textbf{52.17} \\
\hline
\end{tabular}
%\vspace{-3mm}
\end{table}

\begin{figure}[t]
\center
    \begin{tabular}{cc}
        \includegraphics[width=60mm,angle=0,clip]{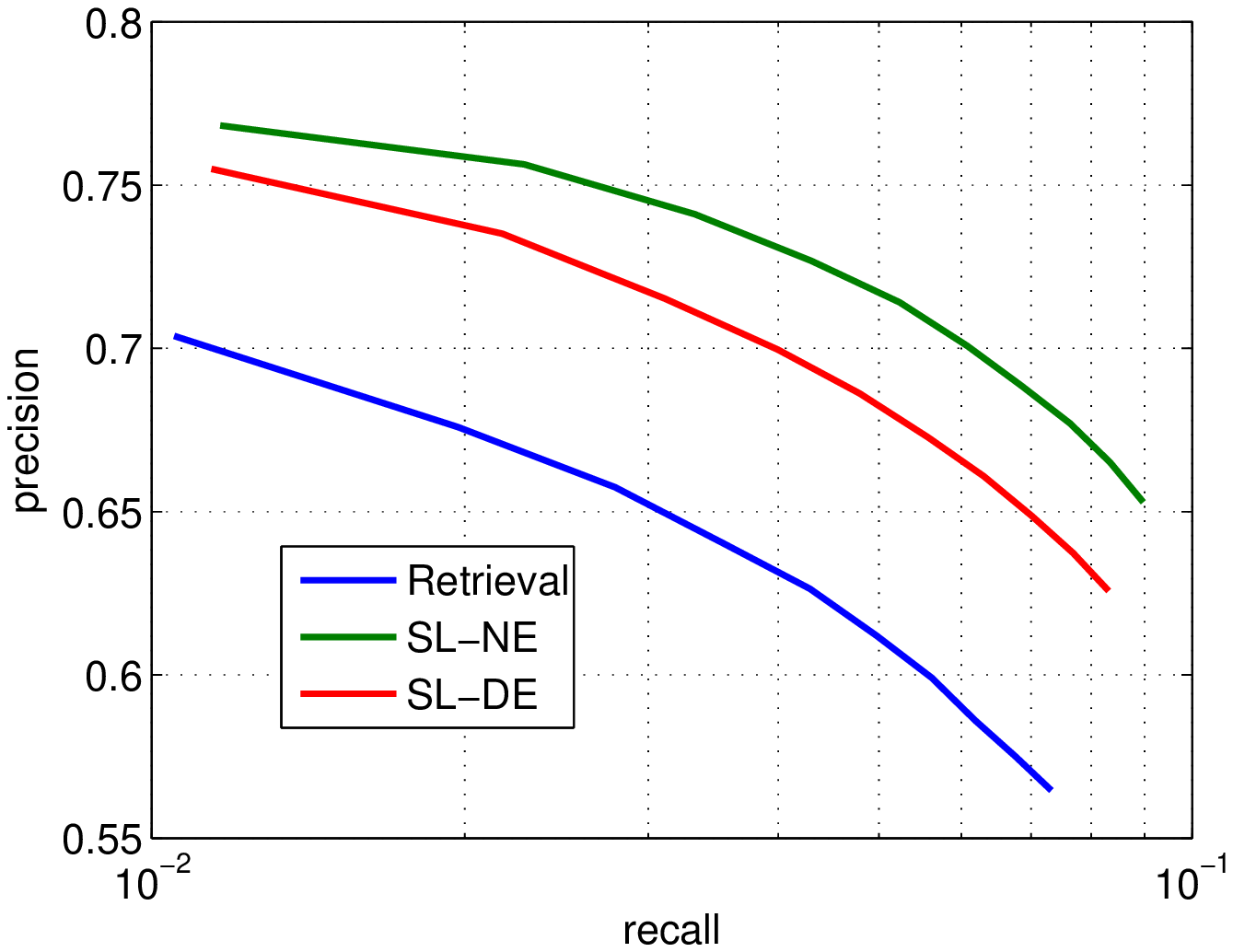} &
        \includegraphics[width=60mm,angle=0,clip]{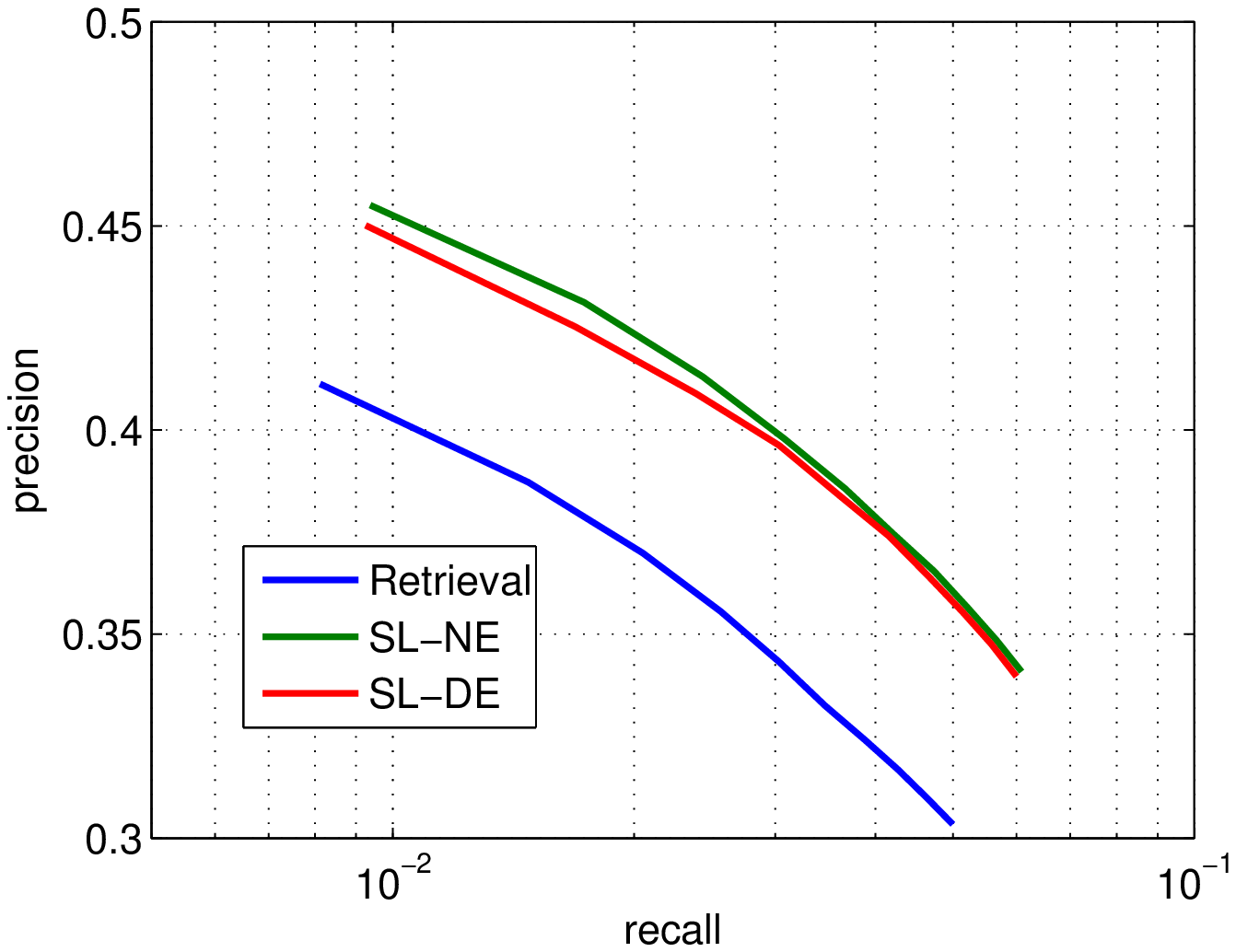} \\
        (a) Caltech 101 & (b) Caltech 256
    \end{tabular}
    \begin{tabular}{c}
        \includegraphics[width=60mm,angle=0,clip]{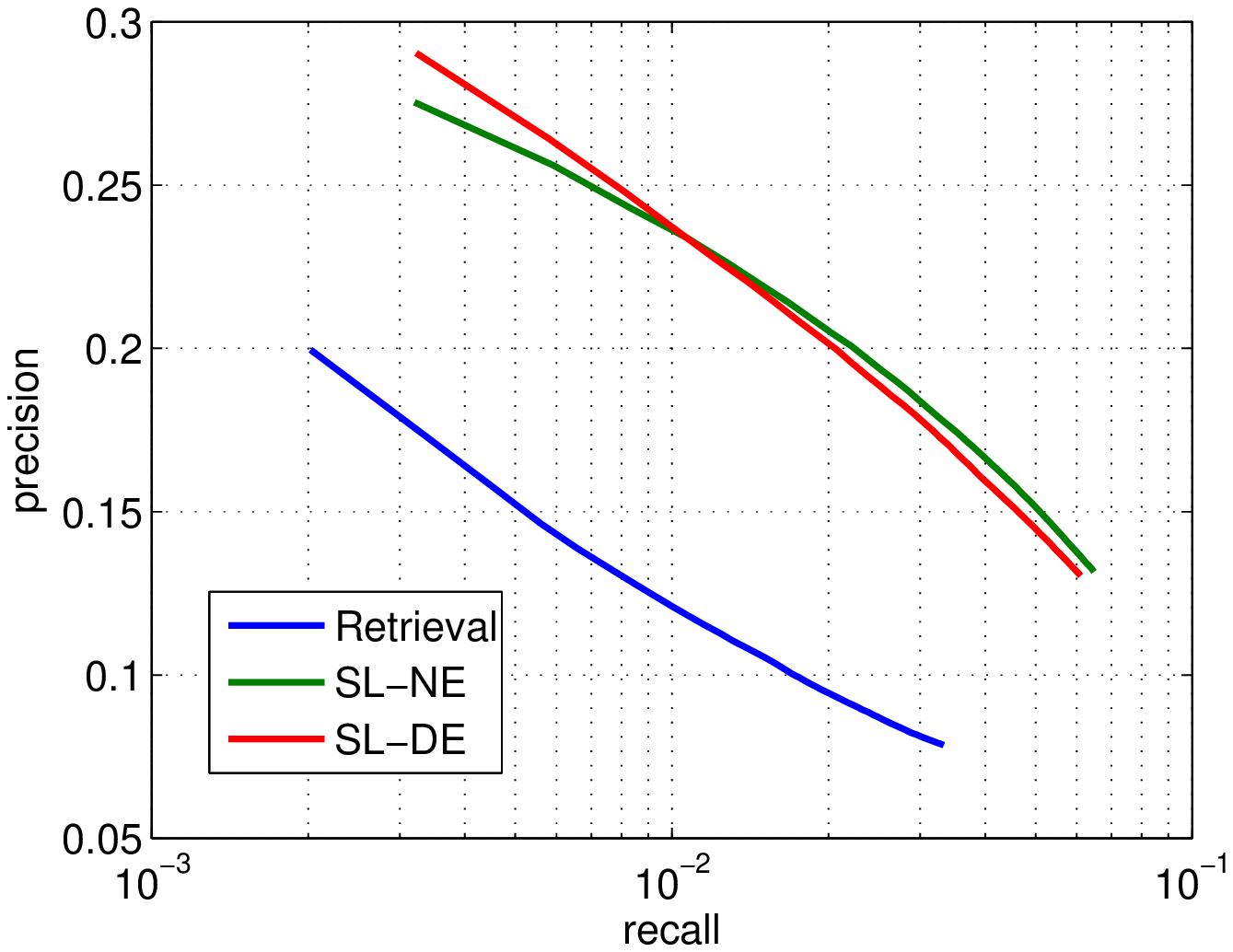} \\ (c) SUN
    \end{tabular}
    \caption{Precision-recall curves for the most similar images found
        by Retrieval \cite{Shen12}, SL-NE and SL-DE.}
\label{fig:PR}
\end{figure}

The similarity functions learned by SL-NE and SL-DE can also be regarded as a rerank function for the similar images found by Retrieval.
Therefore, we also evaluate their retrieval performance by plotting the precision-recall curves in Fig.~\ref{fig:PR}.
SL-NE consistently improves over Retrieval on all the operation points,
and SL-DE achieves very similar performance as SL-NE on the Caltech 256 and SUN data sets.

%30 training, 80 training (minimum 20 testing), 70\% training
%kNN $k$ used: 50, 50, 500
%neighborhood size $|\mathcal{N}|$: 50 50 500
%how many $N$ used for SL-DE: 20 200 100
%what is $d$: 1{,}000 300 500

%ROC
%\begin{figure}[t]
%\center
%    \begin{tabular}{c}
%        \includegraphics[width=80mm,angle=0,clip]{ROC_cal101_comp3} \\ (a) Caltech 101 \\
%        \includegraphics[width=80mm,angle=0,clip]{ROC_cal256_comp3}  \\ (b) Caltech 256 \\
%        \includegraphics[width=80mm,angle=0,clip]{ROC_sun397_comp3}  \\ (c) SUN 397
%    \end{tabular}
%\caption{Two .}
%\label{fig:ROC}
%\end{figure}

%%

\subsection{Results on Large Scale Data}
\label{sec:exp_lrg}

%\begin{figure}[t]
%\center
%    \includegraphics[width=70mm,angle=0,clip]{iteration_ILVRC10_1kPCA_test10}
%\caption{Error rate on validation set versus training iterations on the ILSVRC'10 data using SL-NE.}
%\label{fig:itr_ILSVRC}
%\end{figure}

To further demonstrate the scalability of our methods, we evaluate their performance on
the ImageNet Large Scale Visual Recognition Challenge 2010 (ILSVRC' 10) \cite{Deng09imagenet} data set,
which contains 1.2M training images from 1{,}000 object classes, 50K validation images, and 150K test images.
On this data set, the first 1{,}000 PCA dimensions of LLC features are used as our original feature.
Neighborhood size $|\mathcal{N}|=500$ is used.
%Fig.~\ref{fig:itr_ILSVRC} shows that the online learning algorithm of SL-NE converges very quickly (in 3 iterations)
%on this large data set.
The ensemble similarity functions for SL-DE are trained on 10 evenly block-partitioned PCA subspaces, each of 100 dimension.
The PCA subspaces are preferred here because a small number of them can give a fair result in relatively short time.
%we cannot afford to train similarity functions on a lot of random subspaces for such a big data set.
%A small number of PCA subspaces can give an acceptable performance because they are all uncorrelated.

We compare in Table~\ref{tab:acc_ilsvrc} the top-1 flat classification error of the proposed methods with baseline methods
including linear SVM, Retrieval, two metric learning based approaches \cite{Weinberger09,Mensink12},
and the average performance on the 10 PCA subspaces using SL-NE.
Using the same LLC feature, SL-NE achieves much smaller error rate than Retrieval as well as SVM with an online implementation \cite{BottouSGD},
which implies that similarity learning is more advantageous than one-versus-all
classification models on data sets with a huge number of classes.
Our SL-NE also improves a lot over LMNN, and has a similar performance as Mensink \etal \cite{Mensink12}.
It should be noted that Mensink \etal have used fisher vector \cite{Perronnin10} as their feature representation,
which gives an error rate more than 10\% lower than what LLC achieves when both are used with a SVM classifier.
Therefore, the similarity function learned with SL-NE has made up much of the performance loss due to our weaker feature.
Given a better feature representation, our method has the potential to further reduce the classification error.

\begin{table}[t]
\caption{Top-1 flat error rate (\%) for ILSVRC'10. Results from \cite{Mensink12} are indicated by *.}
\label{tab:acc_ilsvrc}
\centering
\begin{tabular}{C{20mm}C{15mm}C{15mm}C{15mm}C{15mm}C{15mm}C{15mm}C{15mm}}
\hline
SVM	& Retrieval \cite{Shen12} & LMNN \cite{Weinberger09} & Mensink \etal \cite{Mensink12} & PCA subspace & SL-NE  & SL-DE \\
\hline
73.93/60.2* & 77.19 & 72.90* & 65.10* & 80.44 	& 66.37  & 68.00 \\
\hline
\end{tabular}
%\vspace{-5mm}
\end{table}

The SL-DE method has a significant improvement over the similarity function learned in each PCA subspace,
which serves as its building block.
Compared with SL-NE, SL-DE has an error rate less than 2\% higher.
However, it only takes SL-DE 2 hours (excluding the time to retrieve neighborhood, same below)
to train on the 1.2M training set using 10 distributed computers;
and this is much faster than SL-NE which needs almost 2 days to complete the same task.

%converge faster on lower dimension
%but its training time is reduced by approximated 90\% since

\section{Conclusions}
\label{sec:cncl}

A novel image similarity learning method is investigated for better scalability to data set size as well as feature dimensionality.
The similarity function is optimized only for sample pairs within a local neighborhood using large margin neighborhood embedding,
which significantly reduces the number of relative similarity constraints in training
and at the same time enhances the robustness to irrelevant samples.
We also propose the ensemble of similarities for scalability to data dimensionality,
which breaks the high-dimensional problem into several lower-dimensional problems
without much loss in performance.
The proposed method is validated on several image classification data sets, and achieves
competitive accuracies with several existing methods.
More importantly, our approach demonstrates much better scalability
than existing metric learning methods and one-versus-all classifiers.

In future work, we will explore other possibilities to find local neighborhoods for better tradeoff
between search efficiency and accuracy. Potential directions include using hash functions
and joint optimization of neighborhood searching and similarity learning.
%hierarchical (multi-layer) structure
It is also of great interest to investigate efficient learning of discriminative and complementary projection matrices
for ensemble metric learning.
%gain a deeper understanding for the impact of randomly projected similarities
%in classification with theories from low rank matrix approximation and compressive sensing.

%------------------------------------------------------------------------

{%\small
\bibliographystyle{splncs}
\bibliography{ref}
}

\end{document}